\documentclass[10pt,twocolumn,letterpaper]{article}

\usepackage[pagenumbers]{cvpr} % To force page numbers, e.g. for an arXiv version

\usepackage{graphicx}
\usepackage{amsmath}
\usepackage{amssymb}
\usepackage{booktabs}
\usepackage{array}
\usepackage{amsmath}
\usepackage{enumitem}
\usepackage{algorithm}
\usepackage{algpseudocode}
\usepackage{bbm}

\usepackage{epsfig}
\usepackage{graphicx}
\usepackage{amsmath,bm}
\usepackage{amssymb}
\usepackage{bbding}
\usepackage{color}
\usepackage{xcolor}
\usepackage{multirow}
\usepackage{bbm}

\usepackage{sidecap}

\usepackage[accsupp]{axessibility} 
\usepackage[pagebackref,breaklinks,colorlinks]{hyperref}

\usepackage{etoolbox} 
\usepackage[capitalize]{cleveref}
\crefname{section}{Sec.}{Secs.}
\Crefname{section}{Section}{Sections}
\Crefname{table}{Table}{Tables}
\crefname{table}{Tab.}{Tabs.}

\def\cvprPaperID{***} % *** Enter the CVPR Paper ID here
\def\confName{CVPR}
\def\confYear{2022}

\begin{document}

\title{Cross-domain Detection Transformer based on \\ Spatial-aware and Semantic-aware Token Alignment}

\author{Jinhong Deng \space\space\space\space Xiaoyue Zhang \space\space\space\space Wen Li \space\space\space\space Lixin Duan  \\
        % School of Computer Science and Engineering \& Shenzhen Institute for Advanced Study, 
        % \\
        University of Electronic Science and Technology of China\\
      {\tt\small \{jhdeng1997, liwenbnu, lxduan\}@gmail.com, xzhangeo@connect.ust.hk}
      % \and
   }
\maketitle

%%%%%%%%% ABSTRACT
\begin{abstract}
Detection transformers like DETR~\cite{DETR} have recently shown promising performance on many object detection tasks, but the generalization ability of those methods is still quite challenging for cross-domain adaptation scenarios. To address the cross-domain issue, a straightforward way is to perform token alignment with adversarial training in transformers. However, its performance is often unsatisfactory as the tokens in detection transformers are quite diverse and represent different spatial and semantic information. In this paper, we propose a new method called Spatial-aware and Semantic-aware Token Alignment~(SSTA) for cross-domain detection transformers. In particular, we take advantage of the characteristics of cross-attention as used in detection transformer and propose the spatial-aware token alignment~(SpaTA) and the semantic-aware token alignment~(SemTA) strategies to guide the token alignment across domains. For spatial-aware token alignment, we can extract the information from the cross-attention map~(CAM) to align the distribution of tokens according to their attention to object queries. For semantic-aware token alignment, we inject the category information into the cross-attention map and construct domain embedding to guide the learning of a multi-class discriminator so as to model the category relationship and achieve category-level token alignment during the entire adaptation process. We conduct extensive experiments on several widely-used benchmarks, and the results clearly show the effectiveness of our proposed method over existing state-of-the-art baselines.
\end{abstract}

%%%%%%%%% BODY TEXT

\section{Introduction}
\label{sec:intro}

Object detection, as a fundamental task for visual understanding, has been one of the most attractive research problems in the computer vision community~\cite{Faster-RCNN,FCOS,Fast-RCNN,Cascade-RCNN,SSD,YOLO,DETR}. With the thriving of deep convolutional neural networks (CNN)~\cite{AlexNet, Resnet}, many CNN-based object detection approaches (\eg, Faster RCNN~\cite{Faster-RCNN} and FCOS~\cite{FCOS}) have been proposed in the last decade. Recently, detection transformers (\eg, DETR~\cite{DETR}) have gained increasing attention from researchers. Based on the design of visual transformer, detection transformers remove the requirement of hand-designed components such as non-maximum supperssion~(NMS) and anchor generation in traditional CNN-based object detection methods, and at the same time, achieve new state-of-the-art performance in many object detection tasks~\cite{DETR, Deformable-DETR,Sparse-DETR, Conditional-DETR, Anchor-DETR, DINO}. Despite the success of detection transformers, the cross-domain generalization ability remains a challenge when adapting a learned model to a novel domain~(\ie, target domain). Usually, existing detection transformers often suffer from severe performance degradation due to domain discrepancy between the source and target domains~\cite{SFA}.

However, addressing the domain shift issue for detection transformers is non-trivial. Researchers have proposed many ways to improve the cross-domain generalization ability for CNN-based object detectors. For example, a variety of studies for cross-domain object detection (CDOD)~\cite{DA-Faster-RCNN,SWDA,UMT,SSAL,SCDA} are proposed to eliminate the domain discrepancy by aligning the feature distributions of the source and target via adversarial training. Similarly, for the cross-domain detection transformer, a potential and straightforward solution for the cross-domain detection transformer is to perform token alignment with adversarial training, since the visual features are often converted into tokens as the input to the transformer blocks. However, aligning the token distributions is difficult, especially when there exists a significant domain gap between domains.

Recent work~\cite{SFA} attempts to apply adversarial training strategies on tokens in transformers, but the improvements are still unsatisfactory. One of the major reasons is that tokens in detection transformers are quite diverse. In detection transformers~(\eg, DETR), the tokens are passed through several multi-head self-attention layers to obtain new token embeddings for representing different spatial and semantic information. Then, object queries are introduced to probe useful tokens and leverage those tokens to predict the positions and categories of different objects. On the one hand, since some tokens are more useful while less for others, it is desirable to take the importance of tokens into consideration in the cross-domain detection transformer. On the other hand, the semantic information embedded in tokens is also helpful for aligning the token distributions w.r.t. the corresponding category, which can ease the adversarial training process.

In this work, we propose a new cross-domain detection method named Spatial-aware and Semantic-aware Token Alignment~(SSTA) under the transformer framework. In particular, we take advantage of the characteristics of cross-attention as used in the detection transformers and newly developed two strategies, \ie, spatial-aware token alignment~(SpaTA) and semantic-aware token alignment~(SemTA) to guide the token alignment across domains. The cross-attention in the decoder of SSTA utilizes the object queries to aggregate information from encoder outputs (tokens). During this process, only a small part of them are attended to for detecting objects accurately. For spatial-aware token alignment, we can extract the information from the cross-attention map~(CAM) to align the distribution of tokens according to their attention to object queries. For semantic-aware token alignment, we inject the category information into the cross-attention map and construct domain embeddings to guide the learning of a multi-class domain discriminator so as to model the category relationship and achieve category-level alignment during the entire adaptation process. 

We have conducted extensive experiments on three domain adaptive benchmarks, including adverse weather, synthetic-to-real, and scene adaptation, where we achieve new state-of-the-art performance for cross-domain object detection. The experimental results show the effectiveness of our proposed method. We also show the usefulness of each component in our approach by conducting careful ablation studies. The contributions of our work are three-fold: 
\begin{itemize}
\item{We propose a novel approach named Spatial-aware and Semantic-aware Token Alignment~(SSTA) for cross-domain object detection, under the transformer framework. To the best of our knowledge, we make the first attempt to explore the intrinsic cross-attention property for improving the cross-domain generalization ability of detection transformers.}
\item{Two new modules, \ie, token alignment~(SpaTA) and semantic-aware token alignment~(SemTA), are developed respectively to align the token distributions according to their attentions to object queries and to achieve the category-level alignment.}
\item{We conduct extensive experiments on several widely-used benchmarks~(\eg, FoggyCityscapes, Sim10K and BDD100K), and promising results demonstrate the effectiveness of our proposed method over existing state-of-the-art baselines.}
\end{itemize}

%-------------------------------------------------------------------------
\section{Related Work}
\subsection{Object Detection}

Object detection aims to recognize and localize one or multiple objects in a given image. Traditional object detection methods~\cite{Faster-RCNN, FCOS, Fast-RCNN, Cascade-RCNN, SSD, YOLO} are based on convolutional neural networks~(CNN)~\cite{Resnet, VGG,AlexNet} and can be divided into two directions, one-stage, and two-stage methods. Two-stage methods~\cite{Faster-RCNN, Fast-RCNN, Cascade-RCNN} typically first generate some region proposals and then refine their classification and bounding boxes. 
In contrast to two-stage methods, one-stage methods~\cite{FCOS, SSD, YOLO} ignore the proposal generation stage and directly predict the category and coordinates of objects. Although these CNN-based detectors have achieved a remarkable breakthrough, they need many hand-designed components like removing duplicated detections by non-maximum suppression and anchor generation which explicitly encodes our prior knowledge about the task. Recently, Carion \etal, proposed DETR~\cite{DETR} that reaches an end-to-end object detection without anchor generation and any sophisticated post-process procedure. Many DETR-like models~\cite{Deformable-DETR, Sparse-DETR, Conditional-DETR, Anchor-DETR} are proposed to further improve the performance of the DETR model in both convergence speed and accuracy. Among these works, one of the most representative works is Deformable DETR~\cite{Deformable-DETR} which adopts deformable attention mechanism~\cite{deformable-conv} into DETR and designs a multi-scale attention module so that it reduces the training time and improves detection performance significantly. 
% Other works attempt to boost DETR by exploring token sparsity and query initialization. 
Nevertheless, these methods suffer from severe performance degradation due to the domain discrepancy between the training and test domains. To address this problem, we present Spatial-aware and Semantic-aware Token Alignment~(SSTA) to learn domain-invariant token representations. Following \cite{SFA}, we choose Deformable DETR~\cite{Deformable-DETR} as the base detector for a fair comparison.

\subsection{Cross-domain Object Detection}

Cross-domain object detection~(CDOD) aims to transfer the knowledge from the label-rich source domain to the label-scarce target domain by bridging the domain discrepancy between them. Previous works~\cite{PDA, DM, DA-Faster-RCNN, SWDA, SCDA, Mega, UMT, MOTR, SSAL} can be roughly categorized into image translation, self-supervision, and adversarial training. Image translation methods~\cite{PDA, DM} adopt style transfer algorithms to enhance the image diversity so as to reduce the domain gap at the pixel level. Self-supervision approaches~\cite{UMT, MOTR, SSAL, auto-adapt} deploy the pseudo-labeling techniques to provide additional supervision signal for the target domain. Adversarial training methods~\cite{DA-Faster-RCNN, SWDA} align the feature distribution and eliminate the domain discrepancy to bridge the domain gap. Early works align the features with diverse levels, \eg, strong-weak alignment~\cite{SWDA}, global-instance level~\cite{DA-Faster-RCNN}.

However, these methods are based on the Faster RCNN or FCOS, and the transferability of detection transformers remains a challenge. SFA~\cite{SFA} has developed a domain adaptive detection transformer to align domain query feature and token-wise feature and design an additional bipartite matching consistency loss to enhance the feature discriminability. Different from SFA~\cite{SFA}, our SSTA takes advantage of the cross-attention map and leverages the spatial and semantic information to help the token distribution alignment. Our model follows the principle of giving minimal modification to the DETR model so that the inference has no extra overload. To the best of our knowledge, our method is the first domain adaptation work that takes advantage of the characteristics of cross-attention to improve the generalization ability of the DETR model.

\begin{figure*}
% \vspace{-0.3cm}
    \centering
    \includegraphics[width=1.0\linewidth]{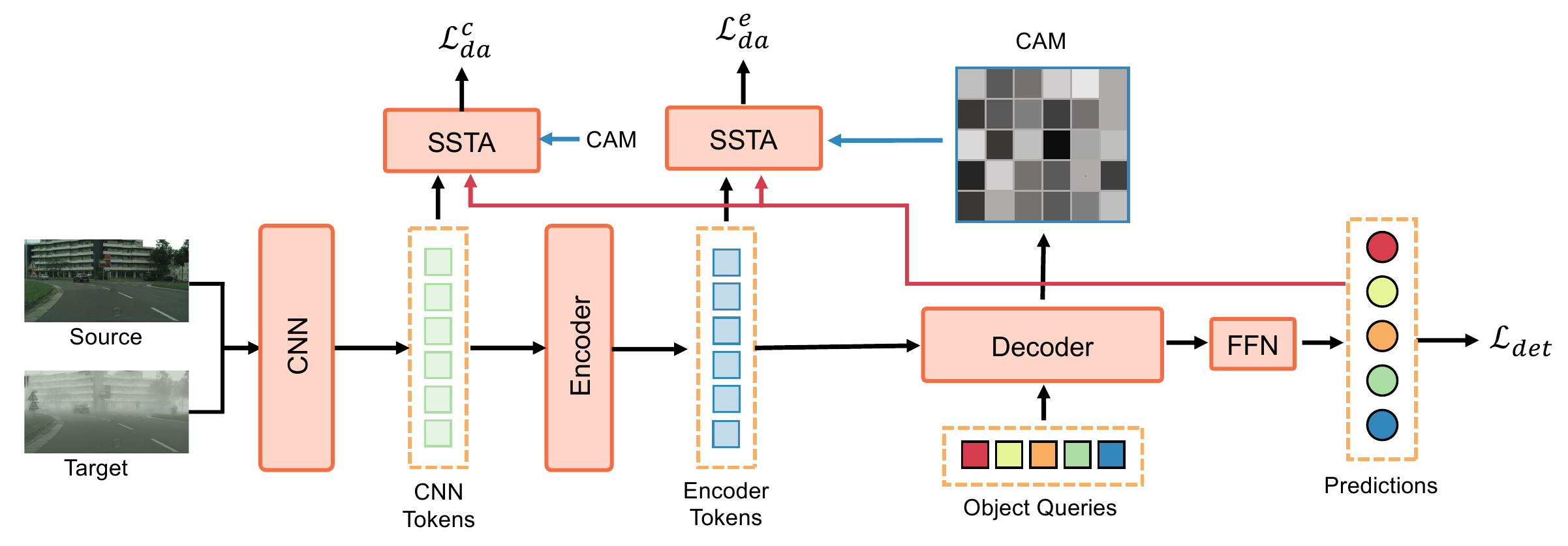}
    \caption{
    The overview of our method. We design a new Spatial-aware and Semantic-aware Token Alignment~(SSTA) module to align CNN token and encoder token distribution across two domains. We take advantage of the characteristics of the cross-attention in the decoder and feed the cross-attention map~(CAM) and the predictions of the detection head~(FFN) to improve the token alignment. The details of the SSTA module are shown in Fig.~\ref{fig:sasata}.}
    \label{fig:framework}
    % \vspace{-5mm}
\end{figure*}

\section{Methodology}

In the task of CDOD, we are given a source domain consisting of labeled images with object bounding boxes and their class labels and a target domain consisting of unlabeled images.
Let us denote $\mathcal{D}_{s} = \{(x_{i}^{s}, \mathbf{y}_{i}^{s})\}_{i=1 }^{N_s}$ drawn from distribution $\mathcal{P}_s$ as the labeled source domain and $\mathcal{D}_{t} = \{x_{j}^{t}\}_{j=1 }^{N_t}$ drawn from distribution $\mathcal{P}_t$ as the unlabeled target domain, where $\mathcal{P}_s \neq \mathcal{P}_t$. And $\mathbf{y}_i^s = \{(\mathbf{b}_{j}^{s}, {c}_{j}^{s})|_{j=1}^m\}$, where $\mathbf{b}_j^s \in \mathbb{R}^{4}$ and ${c}_{j}^{s} \in \{1, \dots, C\}$ are the bounding box and corresponding category for each object, and $m$ is the total number of objects in an image $x_{i}^{s}$. Our goal is to learn an object detection model that performs well on the target domain.

In the following, we introduce the motivation of our proposed method in Sec.~\ref{sec:motivation}. And then, we first give the vanilla token alignment in Sec.~\ref{sec:ta} and describe the detailed design of spatial-aware token alignment (Sec.~\ref{sec:spata}) and semantic-aware token alignment (Sec.~\ref{sec:semta}). Lastly, we give the overall objective of the proposed method.

\subsection{Motivation}
\label{sec:motivation}

In this section, we give a brief preliminary to the DETR model. And then, we demonstrate the cross-domain challenges in DETR as well as our new solution.

\noindent\textbf{DEtection TRansformer~(DETR):} DETR consists of CNN backbone, transformer encoder and transformer decoder. The image $x \in \mathbb{R}^{3 \times H_0 \times W_0}$ are firstly fed into CNN backbone~(\eg, ResNet50~\cite{Resnet}) and to generate a lower-resolution feature map $f \in \mathbb{R}^{C \times H \times W}$, where $C=2048$, $H= {\frac {H_0} {32}}$ and $W= {\frac {W_0} {32}}$. The encoder uses a $1\times 1$ convolution to reduce the channel $C$ into a smaller dimension $d$ and then collapse the spatial dimensions into one dimension, resulting token inputs $z_c \in \mathbb{R}^{d\times N_k}$, where $N_k=WH$ is the length of sequence. The encoder layer adopts tokens $z_c$ along with position embedding to make interaction among tokens and outputs new tokens $z_e \in \mathbb{R}^{d\times N_k}$ through standard architecture that consists of a multi-head self-attention and a feed forward network~(FFN). The decoder comprises of multi-head self-attention and multi-head cross-attention mechanisms. Different with encoder, the decoder first deploys self-attention for $N_q$ object queries and then uses cross-attention~(\ie, encoder-decoder attention) to aggregate features from the outputs of the encoder, resulting a sequence $z_d \in \mathbb{R}^{d\times N_q}$. Finally, the decoder will result $N_q$ predictions. DETR utilizes Hungarian algorithm to find a bipartite matching between the sets of predictions and ground truth. The loss of DETR can be summarized as follows:
\begin{equation}
    \label{eq:det}
     \mathcal{L}_{det} = \mathcal{L}_{cls} + \mathcal{L}_{reg},
\end{equation}
where the $\mathcal{L}_{cls}$ is for classification and $\mathcal{L}_{reg}$ is for bounding boxes regression. 

DETR requires much longer training epochs~(\ie, 500) to converge than traditional detectors and has relatively low detection accuracy on small objects. Thus Deformable DETR~\cite{Deformable-DETR} adopts efficient deformable attention module to replace the dense attention in DETR. The deformable attention mechanism can be naturally extended to aggregating multi-scale features, leading to fast convergence and high performance. Following \cite{SFA}, we choose Deformable DETR~\cite{Deformable-DETR} as the base detector for a fair comparison. For more detail, please refer to \cite{DETR, Deformable-DETR}.

\noindent\textbf{Cross-domain Challenges in DETR:} To improve the generalization ability of detection transformer, a potential solution is to perform token alignment with adversarial learning. Recent work~\cite{SFA} also attempts to apply adversarial training strategies on tokens in transformers, but the improvements are still unsatisfactory. One of the main reasons is that the tokens in detection transformer are quite diverse. In detection transformers~(\eg, DETR), the tokens are passed through several multi-head self-attention layers to obtain new token embeddings for representing different spatial and semantic information. Then, object queries are introduced to probe useful tokens and leverage those tokens to predict the positions and categories of different objects. On the one hand, since some tokens are more useful while less for others, it is desirable to take the importance of tokens into consideration in the cross-domain detection transformer. On the other hand, the semantic information embedded in tokens is also helpful for aligning the token distributions of the corresponding category. This would ease the adversarial training when aligning the token distributions between domains. 

To this end, we propose the spatial-aware token alignment~(SpaTA) and the semantic-aware token alignment~(SemTA) strategies to guide the token alignment across domains by leveraging the characteristics of cross-attention in detection transformer. As shown in Fig.~\ref{fig:framework}, the proposed spatial-aware and the semantic-aware token alignment~(SSTA) module adopts the cross-attention map~(CAM) and predictions of the decoder to align the distributions of tokens from the CNN and encoder. The detail will be presented below.

\subsection{Vanilla Token Alignment}
\label{sec:ta}
Before we dive into the design of our SSTA module, we first introduce the vanilla token alignment. The existing adversarial methods~\cite{DA-Faster-RCNN, SWDA, DA-DETR} usually take a discriminator to reduce domain discrepancy via aligning feature distribution between domains. The discriminator tries to distinguish which domain the features come from, while the feature extractor aims to confuse features and deceive the discriminator in a minimax manner. It can be placed at a certain layer or multiple layers of feature extractor. In practice, a gradient reverse layer~(GRL)~\cite{DANN} is used to connect the discriminator and feature extractor and flips the gradients when it flows through the feature extractor, leading to an end-to-end learning instead of sophisticated multi-stage iterative optimization like~\cite{GAN}. To bridge the domain gap, a naive solution is to simply align the distribution of tokens where the domain discriminator tries to recognize each token. Formally, the adversarial objective of vanilla token alignment can be defined as follows:
\begin{equation}
    \label{eq:baseline_adv}
     \mathcal{L}_{ta} = -\sum_{i=1}^{N_q} \{ d \log(D(z_i)) + (1-d) \log(1-D(z_i)) \},
\end{equation}
where $N_q$ is the length of sequence, $z_i$ is the $i$-th token representation and can be from CNN backbone or transformer encoder, and $d$ is the domain label with $d=1$ for the source and $d=0$ for the target. When the above adversarial learning loss being optimized, the sign of gradient back-propagated from discriminator to feature extractor will be inverted by GRL, thus making the feature extractor learn domain-invariant representations.

The overall objective of vanilla token alignment can be formulated as:
\begin{equation}
    \label{eq:baseline_ol}
    \mathcal{L} = \mathcal{L}_{det} + \lambda\cdot(\mathcal{L}_{ta}^c + \mathcal{L}_{ta}^e),
\end{equation}
where $\lambda$ is the trade-off parameter, and $\mathcal{L}_{ta}^c$ and $\mathcal{L}_{ta}^e$ are the vanilla token alignment loss for the CNN and encoder tokens.

\subsection{Spatial-aware Token Alignment}
\label{sec:spata}

As the analysis in Sec.~\ref{sec:motivation}, object queries are introduced to probe useful tokens and leverage those tokens to predict the positions and categories of different objects. In other words, tokens contribute differently to the detection results. Simply aligning the token distribution between domains has unsatisfactory improvements, as tokens in detection transformer have different importances to object detection task. If we consider the tokens equally contributing to the adversarial training, we will overlook matching the distribution of critical tokens that may contain essential instances and global context for accurately predicting the positions and categories of different objects. Consequently, the efforts to reduce the domain gap will eventually meet difficulties, making the alignment less effective.

\begin{figure*}
\vspace{-0.3cm}

    \centering

    \includegraphics[width=1.0\linewidth]{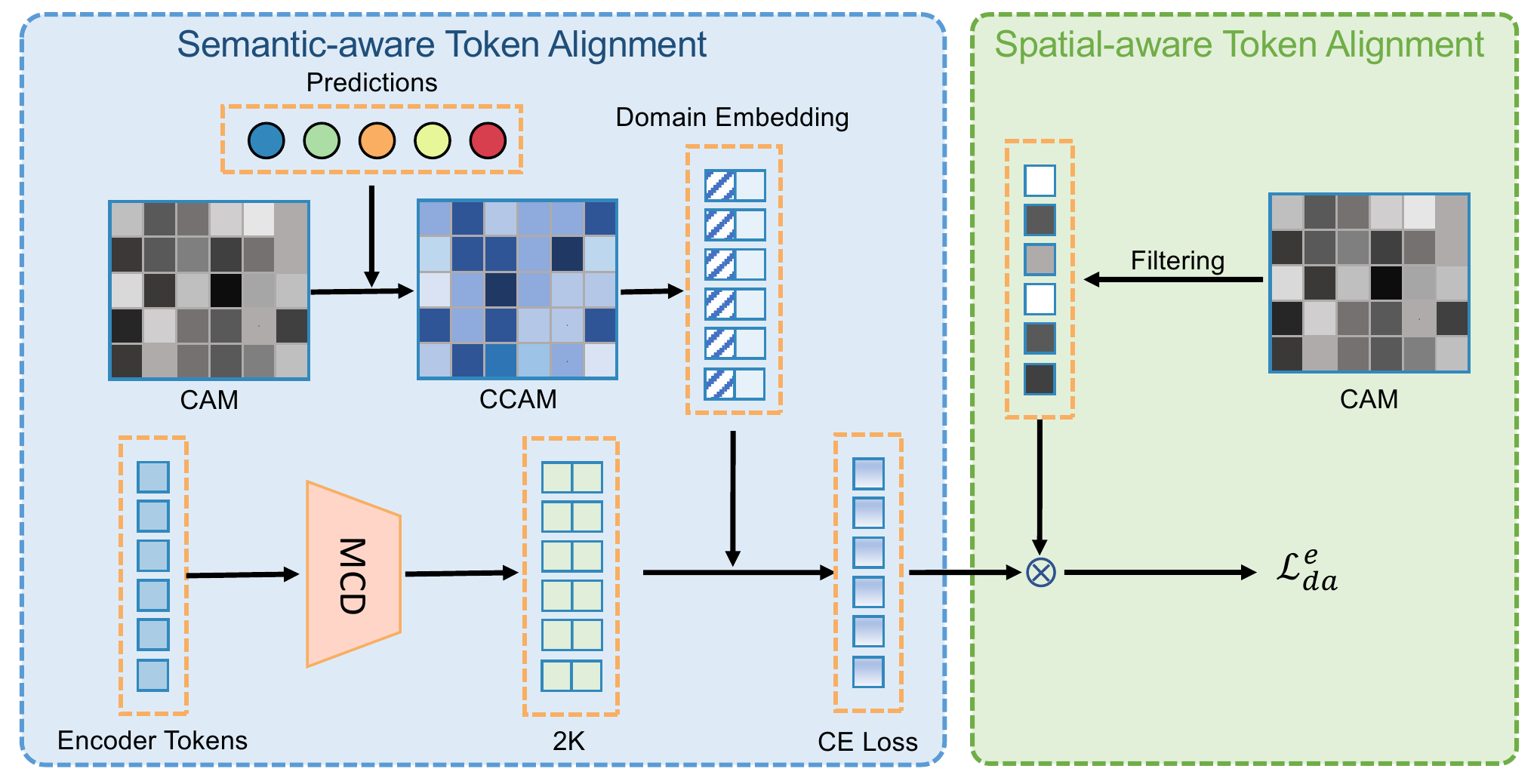}
    \caption{The overview of our Semantic-aware and Spatial-aware Token Alignment~(SSTA) module. Take the SSTA module for encoder tokens as an example. The proposed SSTA module takes the tokens as the input and jointly utilizes Semantic-aware Token Alignment~(SemTA) and Spatial-aware Token Alignment~(SpaTA) to respectively align token distributions. SemTA affiliates the predictions of the detection head into the cross-attention map~(CAM) and obtains a category cross-attention map (CCAM), which can be used to construct domain embedding to guide the learning of a multi-class discriminator~(MCD) to achieve category-level token alignment. The SpaTA utilizes the CAM to give different weights to the adversarial learning of tokens according to their attention to object queries.}
    \label{fig:sasata}
\vspace{-3mm}
\end{figure*}

Motivated by this, we propose a spatial-aware token alignment~(SpaTA) module to discover instance-related tokens and emphasize their alignment by assigning higher weights to these tokens for adversarial training according to their attention to the object queries. Formally, we can obtain the objective as follows:
\begin{equation}
    \label{eq:spa}
     \mathcal{L}_{spa} = \sum_{i=1}^{N_k} (1 + \mathcal{W}^i)\cdot\mathcal{L}_{ta}^{i},
\end{equation}
where $\mathcal{W}^i$ is the weight for $i$-th token, intuitively, the more important the token should be assigned higher weights. As shown in the right part of Fig.~\ref{fig:sasata}, we utilize cross-attention map~(CAM) as an alternative to providing the weights, as object queries probe features by giving different weights to tokens via the cross-attention mechanism.

However, the CAM cannot be directly obtained in deformable attention because of its special design. To this end, the key factor is determining how to obtain the CAM. We scatter and accumulate the cross-attention in the decoder from each object query to discrete token positions in the sequence. The deformable attention applies bilinear interpolation to obtain values from the surrounding position, as attention offset in deformable attention is fractional. Therefore, we also apply bilinear interpolation to obtain CAM. Specifically, let $r$, $\Delta r$, $A$, and $v$ be one of the reference points of the decoder, corresponding offsets, attention weights, and values, respectively. 

For the attention to each token, we can obtain CAM of $i$-th query as follows:
\begin{equation}
    \label{eq:cam}
    %  \sum_{t} A \cdot \mathcal{B}(r, r + \Delta r) \cdot v(t),
     \mathcal{M}_i = {\frac{1}{N_d}} \sum_{l=1}^{N_d} \sum_{(A_l, r, \Delta r)} A_l \cdot \mathcal{B}(t, r + \Delta r),
\end{equation}
where $N_d$ is the number of decoder layer, $\mathcal{B}(\cdot, \cdot)$ is the bilinear interpolation operation, and $t$ enumerates all integral spatial locations of tokens. We provide more details in our Supplementary materials. After obtaining the CAM, we filter out some attentions that are less than a given threshold. 

In summary, the important weight for tokens can be obtained via:
\begin{equation}
    \label{eq:weight}
     \mathcal{W} = \mathcal{M} \odot \mathbbm{1}(\mathcal{M} \geq \tau(\mathcal{M})),
\end{equation}
where $\mathcal{M}$ is the average of CAM for all the queries and $\tau(\mathcal{M}) = mean(\mathcal{M})$ is an adaptive threshold for each sample $x$.

\begin{table*}
  \caption{Average precisions (\%) of different methods on Cityscapes$\rightarrow$FoggyCityscapes.}
  \label{tab:city2foggy}
%   \vspace{-3mm}
  \centering
  \resizebox{0.85\linewidth}{!}{
     \begin{tabular}{c|c|c c c c c c c c|c}
        \hline
        Method & Detector & person & rider & car & truck & bus & train & mcycle & bicycle & mAP \\ \hline
        Faster RCNN~\cite{Faster-RCNN} (Source) & \multirow{9}{*}{\shortstack{Faster\\RCNN}}& 26.9 &38.2 &35.6 &18.3 & 32.4 &9.6 &25.8 &28.6 & 26.9 \\  
        DA-Faster~\cite{DA-Faster-RCNN} & & 25.0 & 31.0 & 40.5 & 22.1  & 35.3 & 20.2  & 20.0  & 27.1  & 27.6  \\  
SWDA~\cite{SWDA} && 29.9 & 42.3 & 43.5 & 24.5  & 36.2 & 32.6  & 30.0  & 35.3  & 34.3 \\   
CFDA     \cite{CFFA}   && 43.2   & 37.4  & 52.1 & 34.7 & 34.0 & 46.9  & 29.9  & 30.8  & 38.6 \\ 
UMT~\cite{UMT}&&33.0&46.7&48.6&34.1&56.5&{46.8}&30.4&37.4 & 41.7  \\
MeGA~\cite{Mega}&& 37.7& 49.0 &52.4 &25.4& 49.2 &46.9&{34.5} &39.0&41.8 \\

ICCR-VDD~\cite{ICCR-VDD} & &33.4& 44.0& 51.7& \textbf{33.9}& \textbf{52.0}& 34.7& 34.2& 36.8& 40.0 \\
ViSGA~\cite{ViSGA} & & 38.8 &45.9 &57.2 &29.9 &50.2 &\textbf{51.9} &31.9 &40.9 &43.3 \\ 
DIDN~\cite{DIDN} & & 38.3 &44.4 &51.8 &28.7 &53.3 &34.7 &32.4 &40.4 &40.5 \\ 
\hline 

FCOS~\cite{FCOS} (Source)  &\multirow{5}{*}{{FCOS}} & 36.9 & 36.3 & 44.1 & 18.6 & 29.3 & 8.4 & 20.3 & 31.9 & 28.2 \\ 
EPM \cite{EPM} &   & 41.9 & 38.7  & 56.7 & 22.6  & 41.5 & 26.8  & 24.6  & 35.5  & 36.0  \\   
SCAN \cite{SCAN}   &    & 41.7 & 43.9 & 57.3 & 28.7 & 48.6 & 48.7 & 31.0 & 37.3 & 42.1 \\  
% SIGMA \cite{hsu2020every}   &&    44.0 & 43.9 & 60.3 & 31.6 & 50.4 & 51.5 & 31.7 & 40.6 & 44.2  \\  \cline{3-11} 
KTNet~\cite{KTNet}& &46.4 &43.2& {60.6}& 25.8& 41.2& 40.4& 30.7& 38.8& 40.9 \\ 
SSAL~\cite{SSAL}& & 45.1 & 47.4 & 59.4 & 24.5 & 50.0  &25.7 & 26.0 & 38.7 & 39.6    \\ 

\hline
Deformable DETR~\cite{Deformable-DETR} (Source) &\multirow{3}{*}{\shortstack{Deformable\\DETR}}& 38.6 & 40.6 & 45.8 & 11.6 & 28.9 & 1.7&18.9&39.1&28.1   \\ 
SFA~\cite{SFA} & & 46.5 &48.6 &62.6 &25.1 &46.2 &29.4 &28.3 &44.0 & 41.3 \\ 
SSTA~(Ours) && \textbf{50.5} & \textbf{53.0} &\textbf{67.2} &24.7 &47.7 &33.0 &\textbf{36.7} &\textbf{46.6} & \textbf{44.9} \\

        \hline
     \end{tabular}
     }
%  \vspace{-3mm}
\end{table*}

\subsection{Semantic-aware Token Alignment}
\label{sec:semta}
Although we have discovered the critical tokens to emphasize their alignment and avoid the influence of noise tokens, the model still has the risk of misalignment during the adaptation process~\cite{Mega,FADA}. The semantic information of tokens is helpful for aligning the token distributions of the corresponding category, so that the model can avoid the class misalignment. For example, the ``car" and the ``truck" instances are forced to be very close in the feature space, deteriorating the model discriminant ability. Therefore, we propose to utilize a multi-class discriminator~\cite{FADA}~(MCD) to capture the category information during adversarial training so that it realizes category-level token alignment. The multi-class discriminator contains not only domain information but also category relationship. Concretely, we remold the single-class discriminator to a multi-classes discriminator that outputs $2K$ logits, where $K = C + 1$, $K$ for the source domain, and others for the target domain. The domain embedding $\textbf{d} \in \mathbb{R}^{2K \times 1}$ of the source and target are $[\mathbf{0} ; \mathbf{s}]$ and $[\mathbf{s} ; \mathbf{0}]$, respectively, where $\mathbf{s} \in \mathbb{R}^{K \times 1}$ is the domain knowledge and $\mathbf{0} \in \mathbb{R}^{K \times 1}$ is all-zero vector. The objective of semantic-aware token alignment can be written as follows:

\begin{equation}
    \label{eq:cda}
    \mathcal{L}_{sem}^i = -\sum_{k=1}^{2K} \mathbf{d}_{k} \cdot \log(\hat{D}(z_i)_{k}),
\end{equation}
where $\hat{D}$ is the multi-class domain discriminator. The key factor is determining how to obtain the domain knowledge $\mathbf{s}$ to build domain embedding for these tokens. As illustrated in the left part of Fig.~\ref{fig:sasata}, we also utilize CAM to extract domain knowledge by injecting the category information into it. In particular, we affiliate the predictions of the detection head into the CAM and obtain a category cross-attention map~(CCAM) which can be formally defined as follows:
\begin{equation}
    \label{eq:ccam}
    \tilde{\mathcal{M}}_k = {\frac{1}{N_{q}^k}} \sum_{i}^{N_q} \mathbbm{1}(\hat{y}_i = k) \cdot \mathcal{M}_i,
\end{equation}
where $\tilde{\mathcal{M}}^k \in \mathbb{R}^{N_k}$ refers to CCAM $\tilde{\mathcal{M}} \in \mathbb{R}^{N_k \times K}$ for category $k$, $N_{q}^k$ is the number of queries that belong to category $k$. The $\hat{y}_i$ is the category prediction from detection head for $i$-th query and $\mathbbm{1}(\cdot)$ is the indicator function where if $\cdot$ is true then equals 1, otherwise 0. The $\mathbf{s}$ can be obtained after apply softmax fuction to the CCAM $\tilde{\mathcal{M}}$. Finally, we can obtain our domain adaptation loss by replacing the $\mathcal{L}_{ta}^i$ by the semantic-aware token alignment in Eq.~\eqref{eq:spa}:
\begin{equation}
    \label{eq:ssta}
    \mathcal{L}_{da} = -\sum_{i=1}^{N_k} (1 + \mathcal{W}^i) \cdot \mathcal{L}_{sem}^{i},
\end{equation}

\subsection{Overall Objective}
\label{sec:overall}

In summary, the overall objective includes the detection loss of Deformable DETR~\cite{Deformable-DETR} on the source domain and domain adaptation loss for the CNN and encoder tokens. In summary, the overall objective can be defined as:
\begin{equation}
    \label{eq:overall_loss}
    %  \mathcal{L}_{D_{c/e}} =  {\over}
    \mathcal{L} = \mathcal{L}_{det} + \lambda\cdot(\mathcal{L}_{da}^c + \mathcal{L}_{da}^e),
\end{equation}
where $\lambda$ is the trade-off parameter, $\mathcal{L}_{da}^c$ and $ \mathcal{L}_{da}^e$ are the domain adaptation loss for the CNN and encoder tokens, respectively.

\section{Experiments}
\label{sec:exp}

Following~\cite{SFA}, we train the model with labeled source data and unlabeled target data and test on the target data. We conduct extensive experiments on three CDOD scenarios. The detection results are evaluated with mean Average Precision~(mAP) under the threshold of $0.5$. 

\subsection{Experimental Setup}
% \subsection{Datasets}
\noindent\textbf{Datasets:} 
Cityscapes dataset was collected for the scenes understanding of road and street. It comprises $2,975$ and $500$ images for training and validation, respectively. It contains $8$ categories: \textit{person, rider, car, truck, bus, train, motorbike}, and \textit{bicycle}. FoggyCityscapes~\cite{FoggyCityscapes} dataset is the foggy version of Cityscapes and generated using the depth information provided by Cityscapes. Thus, it shares the common annotations with Cityscapes. It contains three levels for foggy weather, including $0.01$, $0.15$, and $0.02$. In experiments, we choose the worst foggy weather~(\ie, $0.02$). Sim10K~\cite{SIM10K} dataset is a synthetic dataset rendered by the gaming engine Grand Theft Auto V (GTAV). This dataset contains $10,000$ images with $58,701$ bounding boxes with the category of ``car". BDD100K~\cite{bdd100k} dataset is a large-scale autonomous driving and contains $100$k images with six types of weather, six different scenes, and three categories for the time of day. We extract the subset of daytime, resulting in $36,728$ training and $5,258$ validation images.

Following existing works~\cite{DA-Faster-RCNN,DA-DETR}, we evaluate our method on three benchmark settings: 
\begin{itemize}
    \item{\textbf{Weather Adaptation}: We take Cityscapes as the source domain and FoggyCityscape as the target domain, and the model is trained on the train set of Cityscapes and FoggyCityscape and evaluated on the validation split of FoggyCityscapes.}
    \item{\textbf{Syn2Real}: We explore the adaptation of Sim10K to Cityscapes, we train the model using all the images of Sim10K and the train split of Cityscapes, and report mAP on the validation split of Cityscapes with ``car" category.}
    \item{\textbf{Scene Adaptation}: We use Cityscapes as the source domain dataset and BDD100K containing distinct scenes as a large unlabeled target domain dataset. We evaluate the model on the validation set of BDD100K.}
\end{itemize}

% \subsection{Implementation Details}
\noindent\textbf{Implementation Details:}
Following the default setting in SFA~\cite{SFA}, we adopt Deformable DETR~\cite{Deformable-DETR} as base detector, which contains ResNet-50~\cite{Resnet} backbone pre-trained on ImageNet~\cite{imagenet}, six transformer encoders, six transformer decoders and multiple prediction heads. We adopt Adam~\cite{adam} optimizer to update parameters. For Cityscapes to FoggyCityscapes, we first train the model with a learning rate $2 \times 10^{-4}$ for $40$ epochs, then decay the learning rate to $2 \times 10^{-5}$ for $10$ more epochs. And the trade-off parameter $\lambda$ is set to $1.0$. For Sim10K to Cityscapes and Cityscapes to BDD100K, we set the initial learning rate and the trade-off parameter $\lambda$ to $5 \times 10^{-5}$ and $0.01$ respectively. We pre-train models on source data to obtain reliable CAM. All the experiments are conducted using four V100 GPUs with batch size of $16$, \ie, each GPU contains $2$ source images and $2$ target images. We implement our method with the PyTorch deep learning framework. 
The source code of our method will be released soon.
\subsection{Results}

\begin{table}[t]
  \caption{Average precisions (\%) of different methods on SIM10K$\rightarrow$Cityscapes. 
  }
%   \vspace{-2mm}
  \centering
  \resizebox{0.45\textwidth}{!}{
     \begin{tabular}{c | c | c }
        % \toprule[1.5pt]
        \hline
        Method & Detector & AP on Car \\ \hline
        DA-Faster~\cite{DA-Faster-RCNN} &\multirow{8}{*}{{Faster RCNN}} & 39.0  \\ 
        SCDA~\cite{SCDA} & & 43.0  \\  
        SWDA~\cite{SWDA} & & 40.1  \\ 
        MAF~\cite{MAF} & & 41.1  \\ 
        HTCN~\cite{HTCN} & & 42.5  \\ 
        SAP~\cite{SAP} & &44.9  \\ 
        UMT~\cite{UMT} & & 43.1  \\ 
        ViSGA~\cite{ViSGA} & & 49.3 \\ 
        \hline
        EPM~\cite{EPM} & \multirow{4}{*}{{FCOS}} & 49.0  \\ 
        KTNet~\cite{KTNet}& & 50.7  \\ 
        SCAN~\cite{SCAN}   &    & 52.6  \\ 
        SSAL~\cite{SSAL} & & 51.8 \\
        \hline
        Deformable DETR~\cite{Deformable-DETR}(Source) & \multirow{3}{*}{{Deformable DETR}} & 47.4 \\ 
        SFA~\cite{SFA} & & 52.6  \\ 
        SSTA~(Ours) & & \textbf{57.7} \\
        \hline
     \end{tabular}
     }
    % \vspace{-2mm}
  \label{tab:sim10k2city}
\end{table}

\begin{table*}
  \caption{Average precisions (\%) of different methods on Cityscapes $\rightarrow$ BDD100K.
  }
%   \vspace{-3mm}
  \centering
  \resizebox{0.95\linewidth}{!}{
  		\begin{tabular}{c|c|ccccccc|c}
% 		\toprule[1.5pt]
        \hline
		Methods & Detector & person & rider & car & truck & bus & mcycle & bicycle & mAP \\ 
		\hline
        Faster R-CNN (Source) & \multirow{5}{*}{\shortstack{Faster\\RCNN}} & 28.8 & 25.4 & 44.1 & 17.9 & 16.1 & 13.9 & 22.4 & 24.1 \\ 
        DA-Faster~\cite{DA-Faster-RCNN} &  & 28.9 & 27.4 & 44.2 & 19.1 & 18.0 & 14.2 & 22.4 & 24.9 \\
        SWDA~\cite{SWDA} &  & 29.5 & 29.9 & 44.8 & 20.2 & 20.7 & 15.2 & 23.1 & 26.2 \\
        SCDA~\cite{SCDA} &  & 29.3 & 29.2 & 44.4 & 20.3 & 19.6 & 14.8 & 23.2 & 25.8 \\
        % CR-DA~\cite{ECR} &  & 30.8 & 29.0 & 44.8 & 20.5 & 19.8 & 14.1 & 22.8 & 26.0 \\
        ECR~\cite{ECR} &  & 32.8 & 29.3 & 45.8 & \textbf{22.7} & 20.6 & 14.9 & 25.5 & 27.4 \\
        \hline
		FCOS~\cite{FCOS} (Source) & \multirow{2}{*}{{FCOS}} & 38.6 & 24.8 & 54.5 & 17.2 & 16.3 & 15.0 & 18.3 & 26.4 \\
		EPM~\cite{EPM} &  & 39.6 & 26.8 & 55.8 & 18.8 & 19.1 & 14.5 & 20.1 & 27.8 \\
        \hline
        Deformable DETR~\cite{Deformable-DETR} (Source) & \multirow{3}{*}{\shortstack{Deformable\\DETR}} & 38.4 & 27.1 & 56.1 & 14.6 & 12.3 & \textbf{16.3} & 20.7 & 26.5 \\ 
        SFA~\cite{SFA} &  &  \textbf{40.2} & 27.6 & 57.5 & 19.1 &\textbf{23.4} & 15.4 & 19.2 & 28.9 \\ 
        SSTA~(Ours) &  & 39.4 & \textbf{31.9} & \textbf{59.4} & 16.3 & 17.7 & 15.3 & \textbf{26.2} & \textbf{29.5}   \\ 
% 		\bottomrule[1.5pt]
        \hline
     \end{tabular}
     }
  \label{tab:city2bdd}
%   \vspace{-3mm}
\end{table*}

\begin{table*}
%   \small
  \caption{Ablation studies of SSTA on Cityscapes $\rightarrow$ FoggyCityscapes. TA indicates token alignment.}
%   \vspace{-2mm}
    \centering
    % \vspace{-2mm}
     \begin{tabular}{c | c  c  c | c | c}
        \hline
        Method & TA & SpaTA & SemTA & mAP (\%) & $\Delta$ \\ \hline 
        Deformable DETR~\cite{Deformable-DETR} (Source) & - & - & - & 28.1 & -\\ \hline 
        \multirow{3}{*}{{Proposed}} &$\checkmark$  & && 41.3 & 13.2$\uparrow$ \\ \cline{5-6}
         & & $\checkmark$ & & 42.5 & 14.4$\uparrow$ \\ \cline{5-6}
         & &  &$\checkmark$ & 43.9 & 15.8$\uparrow$ \\ \hline 
        SSTA & & $\checkmark$& $\checkmark$& 44.9 & 16.8$\uparrow$ \\

        \hline
     \end{tabular}
    %  }
%   \end{center}
  \label{tab:ablation}
  \vspace{-2mm}
\end{table*}

\begin{table*}[!ht]

  \caption{Average precisions (\%) w.r.t. different values of $\lambda$ on Cityscapes $\rightarrow$ FoggyCityscapes.}
%   \vspace{-2mm}
  \centering
  \begin{tabular}{c | c | c | c | c | c | c}
    % \toprule[1.5pt]
    \hline
    $\lambda$ & 0.0 & 0.1 & 0.5 & 1.0 & 1.5 & 2.0 \\ \hline
    SSTA & 28.1 & 42.1 & 44.3 & 44.9 & 44.9 & 44.6 \\
    % \bottomrule[1.5pt]
    \hline
  \end{tabular}
  \label{tab:para}
%   \vspace{-3mm}
\end{table*}

We conduct extensive experiments and validate the effectiveness of our method by comparing various state-of-the-art CDOD methods, mainly including three kinds of methods: 1) two-stage detector Faster RCNN 2) one-stage detector FCOS, 3) Deformable DETR. For all the methods, we report the results from the original papers. To validate the effectiveness of our proposed method, we also report the results of the Source model where the model is only trained on the source domain and directly evaluated on the target domain.

\noindent\textbf{Weather Adaptation~(Cityscapes $\rightarrow$ FoggyCityscapes)}: We show the adaptation results in Table~\ref{tab:city2foggy}. We can observe that our proposed method outperforms the previous state-of-the-art approaches by a large margin, reaching $44.9\%$ in terms of mAP. Specifically, Deformable DETR~(Source) achieves $28.1\%$ in terms of mAP, which shows that Deformable DETR has a decent generalization but still suffers from the distribution discrepancy across domains. Both SFA~\cite{SFA} and our SSTA improve the Source baseline. However, our SSTA improved by $3.6\%$ in terms of mAP compared with the counterpart SFA~\cite{SFA}. This demonstrates that our method by leveraging intrinsic cross-attention to conduct spatial-aware and semantic-aware token alignment can effectively improve the generalization ability of detection transformer on the target domain. 

\noindent\textbf{Syn2Real~(Sim10K $\rightarrow$ Cityscapes)}: The results of synthetic-to-real adaptation are presented in Table~\ref{tab:sim10k2city}. Our proposed method SSTA reaches the highest mAP~($57.7\%$) that exceeds all compared state-of-the-art methods, including the two-stage, one-stage, and DETR works, by a large margin, that is $5.1\%$ in terms of mAP over best-performing one-stage detector SCAN~\cite{SCAN} and DETR counterpart SFA~\cite{SFA}. These results verify the effectiveness of our SSTA.

\noindent\textbf{Scene Adaptation~(Cityscapes $\rightarrow$ BDD100K)}: The quantitative results are shown in Table~\ref{tab:city2bdd}. According to Table~\ref{tab:city2bdd}, our method SSTA achieves the new state-of-the-art results of $29.5\%$ in terms of mAP, which surpasses the previous works. This again demonstrates the generalization of our method.

% \section{Discussion}
\noindent\textbf{Ablation Studies:} To further verify the effectiveness of our proposed method, we have conducted detailed ablation studies by isolating each component of our SSTA. The experimental results are shown in Table~\ref{tab:ablation}. In particular, our SpaTA significantly boosts the baseline, leading to $14.4\%$ mAP improvements compared with Source model~($28.1\%$). This implies that the CAM can provide sufficient information to discover critical tokens, and emphasizing their contributions to distribution alignment will significantly improve the generalization ability of Deformable DETR. Moreover, SemTA also improves the accuracy of Deformable DETR, achieving $43.9\%$ in terms of mAP. These improvements mainly come from our SemTA considering category information during token alignment and thus avoiding class misalignment. By synergizing SpaTA and SemTA together, we obtain $44.9\%$ in terms of mAP, which shows their complementary to each other.

\noindent\textbf{Parameter Analysis:} We also investigate the influence of the trade-off parameter $\lambda$ which is used to balance the weight between the source detection loss $\mathcal{L}_{det}$ and the domain adaptation loss. Table~\ref{tab:para} summarizes the experimental results on Cityscapes $\rightarrow$ FoggyCityscapes. Note that when $\lambda = 0$, the method degenerates to the Source model. According to Table~\ref{tab:para}, we can conclude that our proposed SSTA consistently improve the generalization ability of Deformable DETR in a wide range of $\lambda$, and $\lambda = 1.0$ and $\lambda = 1.5$ are the bests among them.

\section{Conclusion}
Detection transformers (\eg, DETR) have shown promising results for object detection, when training and test images come from the same domain. However, they usually do not work well for cross-domain problems. In this work, we tackle cross-domain object detection by proposing a novel approach named Semantic-aware and Spatial-aware Token Alignment~(SSTA) under the transformer framework. In SSTA, two new modules \ie, spatial-aware token alignment~(SpaTA) and semantic-aware token alignment~(SemTA), are developed to guide the token alignment across domains. Promising results on benchmark datasets demonstrate the effectiveness of our method.

\noindent\textbf{Limitation:} Although our method outperforms existing cross-domain object detection works, it still faces challenges in detecting objects of rare classes. For example, the ``truck'' and ``train'' classes in Table~\ref{tab:city2foggy} have relatively low AP compared with other classes (\eg, ``car''). We conjecture that this is caused by the label shift between the source and target domains. In the future, we will study how to improve the detection performance of our SSTA for these classes.

%%%%%%%%%%%%%%%%%%%%%%%%%%%%%%%%%%%%%%%%%%%%%%%%%%%%%%%%%%%%
%%%%%%%%%%%%%%%%%%%%%%%%%%%%%%%%%%%%%%%%%%%%%%%%%%%%%%%%%%%%
%%%%%%%%%%%%%%%%%%%%%%%%%%%%%%%%%%%%%%%%%%%%%%%%%%%%%%%%%%%%
%%%%% \paragraph{Acknowledgement} Thanks for antinomy!%%%%%%
%%%%%%%%%%%%%%%%%%%%%%%%%%%%%%%%%%%%%%%%%%%%%%%%%%%%%%%%%%%%
%%%%%%%%%%%%%%%%%%%%%%%%%%%%%%%%%%%%%%%%%%%%%%%%%%%%%%%%%%%%
%%%%%%%%%%%%%%%%%%%%%%%%%%%%%%%%%%%%%%%%%%%%%%%%%%%%%%%%%%%%

%%%%%%%%% REFERENCES
{\small
\bibliographystyle{ieee_fullname}
\bibliography{egbib}
}

\end{document}